\documentclass[preprint,5p,lefttitle]{elsarticle}

\usepackage{microtype}
\usepackage{amsmath}
\usepackage{amsfonts}
\usepackage{booktabs}
\usepackage{colortbl,multirow,multicol}
\usepackage{tabularx}
\usepackage{threeparttable}
\usepackage[edges]{forest}
\usepackage[hyperfootnotes=true, hidelinks]{hyperref}
\usepackage{dblfloatfix}
\usepackage{caption}
\usepackage{makecell}
\usepackage{svg}
\begin{document}

\newcommand{\revisioncolor}{black}

\begin{frontmatter} 


\title{\textbf{Open-world Story Generation with Structured Knowledge Enhancement: A Comprehensive Survey}}

\author[NJU]{Yuxin Wang\fnref{fn1}}
\ead{yuxinwangcs@outlook.com}

\address[NJU]{{State Key Laboratory for Novel Software Technology, Nanjing University}, Nanjing, 210023, Jiangsu, China}

\author[HIT]{Jieru Lin}
\ead{v-jierulin@microsoft.com}


\address[HIT]{Harbin Institute of Technology, Harbin, 150001, Heilongjiang, China}

\author[MSRA]{Zhiwei Yu}
\ead{zhiwyu@microsoft.com}

\author[NJU]{Wei Hu\corref{cor1}}
\ead{whu@nju.edu.cn}

\author[BAAI]{Börje F.~Karlsson\fnref{fn1}\corref{cor1}}
\ead{borje@baai.ac.cn}

\address[MSRA]{Microsoft Research Asia, Beijing, 100080, Beijing, China}
\address[BAAI]{Beijing Academy of Artificial Intelligence, 100086, Beijing, China}
\fntext[fn1]{This work was performed while the authors were at Microsoft Research Asia.}

\cortext[cor1]{Corresponding authors.}

\begin{abstract}
    Storytelling and narrative are fundamental to human experience, intertwined with our social and cultural engagement.
    As such, researchers have long attempted to create systems that can generate stories automatically.
	In recent years, powered by deep learning and massive data resources, automatic story generation has shown significant advances.
	However, considerable challenges, like the need for global coherence in generated stories, still hamper generative models from reaching the same storytelling ability as human narrators. 
	To tackle these challenges, many studies seek to inject structured knowledge into the generation process, which is referred to as structured knowledge-enhanced story generation. 
	Incorporating external knowledge can enhance the logical coherence among story events, achieve better knowledge grounding, and alleviate over-generalization and repetition problems in stories.
	This survey provides the latest and comprehensive review of this research field:
	(i) we present a \textcolor{\revisioncolor}{systematic} taxonomy regarding how existing methods integrate structured knowledge into story generation;
	(ii) we summarize involved story corpora, structured knowledge datasets, and evaluation metrics;
	(iii) we give multidimensional insights into the challenges of knowledge-enhanced story generation and cast light on promising directions for future study.

\end{abstract}


\end{frontmatter}

\section{Introduction} \label{sec_intro}

Humans think in terms of stories, the world is understood in terms of stories, and
people often approach problem-solving and new ideas by referencing stories they
already understand \cite{Schank_1995}. 
Storytelling and narrative are therefore fundamental to human experience, intertwined with our social and cultural engagement \cite{polletta2011sociology}. 
Not only do people spread, preserve, and revise stories for informational, educational, or entertaining purposes \cite{mclellan2006corporate,abrahamson1998storytelling,jenkins2014transmedia}, but the
human brain has a natural affinity for
creating them \cite{wallis:2007}.
\textcolor{\revisioncolor}{As such, \textit{story generation} has consistently drawn attention from the natural language processing (NLP) community \cite{Alhussain2021Survey}.}
Beyond simple text generation, automatic story generation concerns spontaneously generating a coherent narrative, often prompted by a context or goal \cite{Great_Expectations}.


\textcolor{\revisioncolor}{
Early efforts on automatic story generation were generally based on formalisms or symbolic approaches \cite{mcc3609_on_the_craft,Fabula,icec14_conceptual_model}.
Multiple storytelling systems were developed along these lines to assist authors focusing on specific scenarios or to enable interactive storytelling. Some significant systems include: Tailspin~\cite{meehan1977tale} and Minstrel~\cite{turner1993minstrel} among the pioneering systems, DEFACTO\cite{sgouros1999dynamic}, Mimesis~\cite{young2001overview}, IDtension \cite{Szilas}, and Fa\c{c}ade \cite{mateas2005structuring} for interactive drama, GADIN~\cite{barber2007dynamic} and IDA~\cite{magerko2005story} for later author-centric approaches, and Suspenser \cite{cheong2014suspenser} for suspense novel writing.
The majority of these works targetted \textit{close-world} stories, where all the story elements, such as characters, emotions, places, and events, are constrained by pre-defined world settings.
}
A significant example of formalism that inspired such approach is Propp's study of folktales \cite{Propp}, which summarizes the composition of tales and patterns of event development.
Though these symbolism- and formalism-based methods can automatically generate stories with little human intervention, they require arduous knowledge engineering to build word settings for specific domains.


With the emergence of deep learning algorithms and the increased capabilities of generative models, efforts started to shift to neural story generation.
These models can produce text sequences by modeling the probability distribution of word sequences (i.e. language model) \cite{martin2017improvisational}.
This process requires no pre-defined world settings and is referred to as \textit{open-world} story generation.
Later, based on the Transformer architecture \cite{Transformer}, different efforts begin to build large pre-trained language models (PLMs) by feeding them with massive amounts of unsupervised data. 

With such massive training data and huge parameter spaces, PLMs show a great ability to capture syntactic and semantic information within text \cite{GPT}.
They can produce grammatically correct and semantically intelligible short stories, even sometimes surprise readers.
Recent trends on story generation tend to be based on PLMs and conduct specific manipulations to cater to different scenarios and requirements, such as persona control \cite{EmoRL,CONPER,mori2022computational}, content planning \cite{PlotMachine,Aristotelian_Rescoring,Plug-Play,tan-etal-2021-progressive}, human-AI collaboration \cite{nichols2020collaborative,StoryBuddy,Wordcraft}, and explainability \cite{Explainable}. 

However, there are still many known problems and open issues in leveraging PLMs to generate stories.
One major challenging issue is the lack of logical coherence and consistency \cite{see-etal-2019-massively,Common-Posttrain}, especially in long text. PLMs have shown unsatisfactory performance on capturing the semantic dependency among sentences and events. Generated events can be irrelevant, contradictory, and logically ``out-of-order''. The personas and emotions of protagonists can be ``out-of-character''; and storylines can deviate from the theme or genre.

Another open problem is repetition \cite{see-etal-2019-massively,shirai2021neural}. PLMs are still likely to get stuck in a state which keeps producing repetitive token sequences. 
Moreover, PLMs show deficient knowledge grounding \cite{zhao-etal-2020-knowledge-grounded}. As a result, generated stories can be bland, not informative and specific.

Furthermore, story generation via PLMs lack controllability \cite{lin-riedl-2021-plug}. It is hard to control the content produced by PLMs beyond the given prompt, such as enforcing or editing the narrative style, character persona, specific events, or story topic. 

To alleviate these problems, many recent studies begin to seek external knowledge to enhance story generation (introduced in Section~\ref{sec_method}).
The intuition being that, by integrating structured knowledge that encodes explicit facts or rules into the generation process, such  logical rules and inter-dependencies help to make generated stories more coherent and consistent.
In addition, learning from knowledge-intensive data helps to achieve better knowledge grounding.
Structured knowledge enhancement has shown to be effective in lifting the quality of generated stories \cite{Common-Posttrain,MKR,Coep}.

There have been a few recent surveys on text and story generation in general. But these don't cover state-of-the-art developments in using structured knowledge to improve story generation.
Hou et al. \cite{Hou-Survey} give an overview of early deep learning-based works that train seq2seq models from scratch. They classify these works into three categories according to the user constraints: theme-oriented, storyline-oriented, and human-machine interaction-oriented.
Alabdulkarim et al. \cite{Alabdulkarim-Survey} focus on the challenges of PLM-based story generation models and specific approaches to address them. They highlight the challenges of controllability and commonsense.
Alhussain et al. \cite{Alhussain2021Survey} gives a more thorough development of story generation, especially early symbolism- and formalism-based works. Despite its somewhat comprehensive scope, it does not elaborate on recent structured knowledge-enhanced works nor presents a taxonomy. 
Lastly, Yu et al. \cite{Yu-Survey} summarize the designs and techniques used in recent knowledge-enhanced text generation literature. However, they cover a broad spectrum including dialogue systems, question answering, and summarization, while lacking an organized and insightful discussion on story generation.

Recent years have witnessed big progress in structured knowledge-enhanced story generation in terms of scenario, methodology, data sources, and story quality measurements.
A systematical and insightful overview of existing methods is necessary for advancing the field.
Yet, there are still many challenges and bottlenecks requiring perceptive analysis and specific solutions.

Our survey serves a latest and comprehensive view of structured knowledge-enhance story generation to fill the gaps of previous studies:
\begin{itemize}
    \item We summarize relevant works and present a methodological taxonomy regarding how they integrate structured knowledge. Including details of non-trivial operations involved in knowledge utilization.
    \item We discuss available story corpora and structured knowledge datasets, as well as evaluation metrics used in story generation, along with a discussion on their characteristics from a practical point of view.
    \item We analyze the limitations and challenges of existing implementations, datasets, and evaluation metrics. Meanwhile, attempting to point light on future directions.
\end{itemize}

The rest of this paper is organized as follows:
Section~\ref{sec_terminology} presents the definitions of relevant and necessary concepts regarding stories, story generation, and structured knowledge. 
In Section~\ref{sec_method}, we present a taxonomy of existing works.
Then, in Section~\ref{sec_operation}, we discuss non-trivial aspects involved in integrating and leveraging structured knowledge in storytelling.
Section~\ref{sec_datasets} summarizes the available story corpora and structured knowledge datasets and presents their detailed statistics and characteristics.
Section~\ref{sec_eval} presents evaluation metrics applied in story generation.
In Section~\ref{sec_limitation}, we give insights into the current limitations and challenges from multiple aspects. Followed by a discussion on 
prospects for promising future research directions.
Finally, Section~\ref{sec_conclusion} concludes this survey.
\section{Terminology Definition} \label{sec_terminology}

\begin{figure}[t]
\includegraphics[width=\columnwidth]{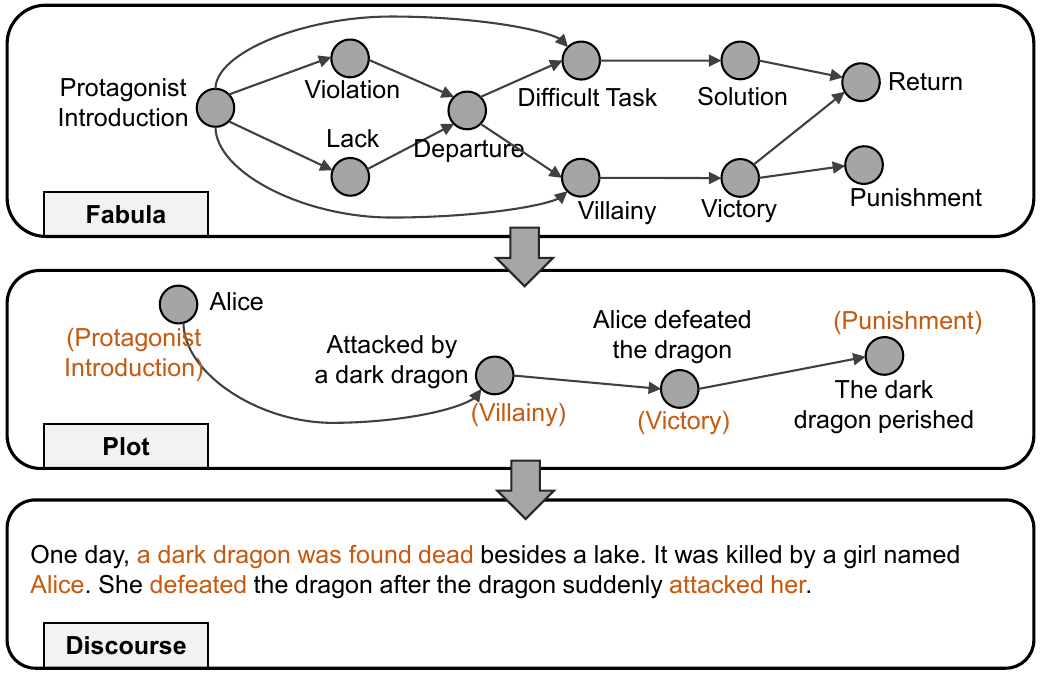}
\centering
\caption{Three layers of narratological concepts about story: fabula, plot, and discourse.}
\label{fig:fabula}
\end{figure}

\subsection{Three Conceptual Layers about Story}
The terminology used to talk about stories in related literature, such as ``narrative'' and ``discourse'' is usually used ambiguously and hard to clarify without a specific context.
Accurately understanding these terms lays the foundation for comprehending and properly discussing related works and analyzing story structure.
Here we present their definitions from the narratological standpoint (adapted to use clearly distinct terms not focused only on text). 
There are three conceptual levels regarding stories: \textit{fabula}, \textit{plot}, and \textit{discourse} \cite{Narratology,Fabula}. 
Figure~\ref{fig:fabula} provides a top-down illustration of them.
\begin{itemize}
    \item \textbf{Fabula}. Fabula defines the space of possible events in a story world and their logical order \cite{castricato-etal-2021-fabula}. It can be seen as an intricate causal network \cite{Fabula}. Besides, it regulates story elements such as places and characters.
    \item \textbf{Plot}. A plot consists of a specific chain of events in a fabula and its constituent elements. It is self-contained, logically coherent, and thematically consistent. Given the same fabula, many different plots can take place. 
    \item \textbf{Discourse}. One story can be presented to readers with various narrative manipulations or media. Discourse\footnote{In the studies by Mieke \cite{Narratology} and Swartjes \cite{Fabula}, they call the  \textit{discourse} level \textit{narrative text} and \textit{presentation} layer. } is a particular way of storytelling \cite{white1990content}. It can completely change the effect of a story on readers. This level is what readers directly interface with. 
\end{itemize}

\textcolor{\revisioncolor}{
With the rise of deep learning and large language models, which empowers a more open-world approach and less restricted scenarios, the \textit{fabula} layer has been much less often targeted in more recent automatic story generation literature.}
However, it remains important in studies requiring meticulous constraints and order of plots \cite{cantoni2020procedural,Ware_fabula,santos2022changing}.
\textcolor{\revisioncolor}{
Regarding the story generation task, most present works focus on generating text at the \textit{discourse} level, while some of them touch the \textit{plot} level, usually involving a top-down hierarchical generation process.
Generally, the most common generation tasks fall into four scenarios: 1) starting sentence-prompted generation, generating subsequent text for a given story beginning; 2) story ending generation, producing a reasonable ending for a given story context; 3) keyword-prompted generation, generating a story conditioned upon a list of keywords, key-phrases, or key plots; and 4) plot infilling, generating coherent missing story pieces between given plot events.
Table~\ref{table:task} gives examples of each of the scenarios.
}

\begin{table}[!t]
\centering
\footnotesize
    \resizebox{\columnwidth}{!}{
    \begin{tabularx}{\linewidth}{X}
    \toprule
    \textbf{Scenario: Starting sentence-prompted generation}  \\
    \midrule
    \textbf{Prompt:} \textit{Rita saw a wounded bird.} \newline 
    \textbf{Output:} \textit{Her heart sank at the sight of its frailty. She cradled it tenderly and rushed to a nearest wildlife center.}\\
    \midrule
    \textbf{Scenario: Story ending generation} \\
    \midrule
    \textbf{Prompt:} \textit{Tom had a puppy, but the potty training spot was untouched. A bad smell filled the room. \underline{~~~~~~~~}} \newline 
    \textbf{Output:} \textit{He found dog poop under the sofa.} \\
    \midrule
    \textbf{Scenario: Keyword-prompted generation} \\
    \midrule
    \textbf{Prompt:} \textit{[cat, black bird, big tree, jump]} \newline 
    \textbf{Output:} \textit{A cat spotted a black bird on a big tree. With a daring jump, it tried to catch the elusive bird, but it danced out of reach, leaving the determined cat chasing the skies.} \\
    \midrule
    \textbf{Scenario: Plot infilling} \\
    \midrule
    \textbf{Prompt:} \textit{Jack woke up late this morning. \underline{~~~~~~~~} He feels much better now.} \newline
    \textbf{Output:} \textit{He realized he had forgotten to set his alarm. Then, he took a refreshing shower and brewed a strong cup of coffee.} \\
    \bottomrule
    \end{tabularx}}
    \caption{\textcolor{\revisioncolor}{Examples of four scenarios covered by story generation tasks.}}
    \label{table:task}
\end{table}
 
\subsection{Generative Language Model}
Story generation models commonly capture and mimic word dependencies of input text to generate story continuances.
A language model is the probability distribution over sequences of words, guiding how generative models predict and produce new sequences \cite{Mnih_languagemodel}.
Given an input sequence $X=\{x_1, x_2, ..., x_n\}$, generative models aim to maximize the likelihood of generating the target sequence $Y=\{y_1, y_2, ..., y_m\}$. 
\textcolor{\revisioncolor}{Here, each $x_i$ and $y_i$ is a word token.}
Story generation generally applies an autoregressive language model for left-to-right generation flow, of which the learning objective $\mathcal{L}_S$ is calculated as the negative likelihood of cumulative probability of words in $Y$:
\begin{equation}\label{eq:lm_loss}
    \mathcal{L}_S = - \sum_{t=1}^{m} \log P(y_t|y_{<t}, X, \boldsymbol{\theta}),
\end{equation}
where $y_{<t}$ denotes the words before the $t$-th step, $\boldsymbol{\theta}$ are the trainable parameters, and $m$ is the target sequence length.

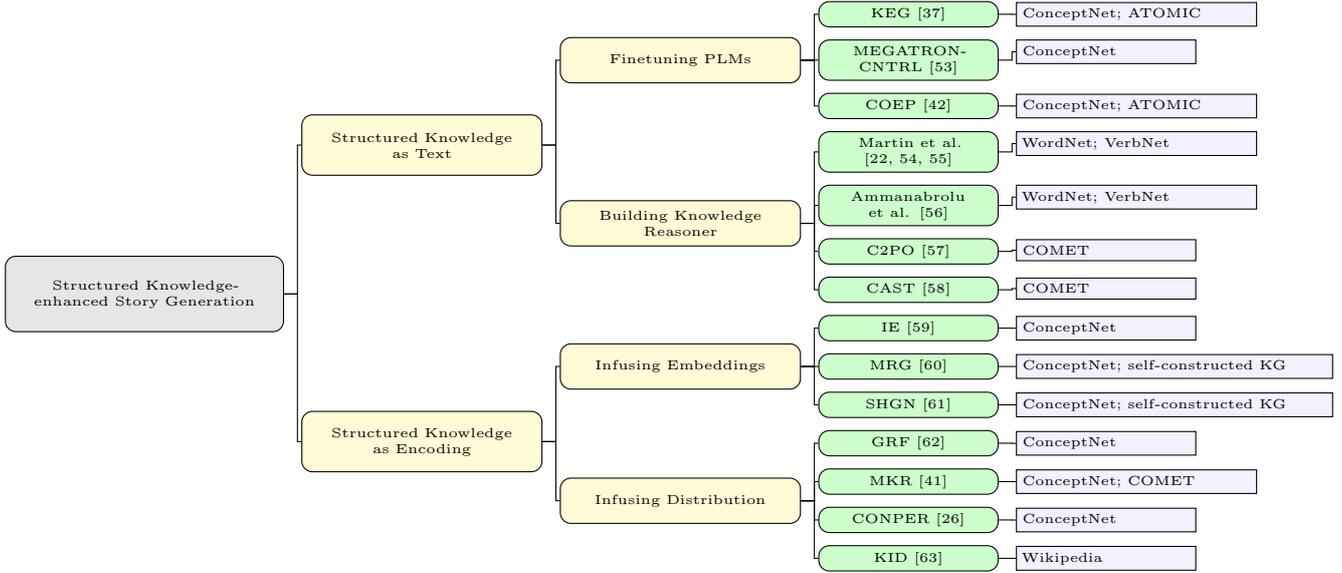
\begin{figure*}[!t]
	\tiny
	\centering
	\begin{forest}
		for tree={
			forked edges,
			grow'=0,
			draw,
			rounded corners,
			node options={align=center},
		},
		[Structured Knowledge-enhanced Story Generation, text width=3.5cm, minimum height=10mm, fill=black!10
			[Structured Knowledge \\ as Text, text width=3cm, minimum height=8mm, for tree={fill=yellow!20}
				[Finetuning PLMs, text width=3cm, for tree={fill=green!20}, minimum height=6mm, fill=yellow!20
					[
					KEG \cite{Common-Posttrain},
					text width=2.2cm,
					node options={align=center}
					    [ConceptNet; ATOMIC, 
					    text width=3cm, 
					    node options={align=left}, 
					    rounded corners=false,
					    fill=blue!5]
					]
					[
					MEGATRON-CNTRL \cite{MEGATRON},
					text width=2.2cm,
					node options={align=center}
					    [ConceptNet, 
					    text width=2.2cm, 
					    node options={align=left}, 
					    rounded corners=false,
					    fill=blue!5]
					]
					[
					COEP \cite{Coep},
					text width=2.2cm,
					node options={align=center}
					    [ConceptNet; ATOMIC, 
					    text width=3cm, 
					    node options={align=left}, 
					    rounded corners=false,
					    fill=blue!5]
					]
				]
				[Building Knowledge \\ Reasoner, text width=3cm, minimum height=6mm, for tree={fill=green!20}, fill=yellow!20
				    [
					Martin et al. \cite{martin2017improvisational,martin2018event,martin2018dungeons},
					text width=2.2cm,
					node options={align=center}
					    [WordNet; VerbNet, 
					    text width=3cm, 
					    node options={align=left}, 
					    rounded corners=false,
					    fill=blue!5]
					]
					[
					Ammanabrolu et al. \cite{ammanabrolu2019guided},
					text width=2.2cm,
					node options={align=center}
					    [WordNet; VerbNet, 
					    text width=3cm, 
					    node options={align=left}, 
					    rounded corners=false,
					    fill=blue!5]
					]
					[
					C2PO \cite{C2PO},
					text width=2.2cm,
					node options={align=center}
					    [COMET, 
					    text width=2.2cm, 
					    node options={align=left}, 
					    rounded corners=false,
					    fill=blue!5]
					]
					[
					CAST \cite{CAST},
					text width=2.2cm,
					node options={align=center}
					    [COMET, 
					    text width=2.2cm, 
					    node options={align=left}, 
					    rounded corners=false,
					    fill=blue!5]
					]
				]
			]
			[Structured Knowledge \\ as Encoding, text width=3cm, minimum height=8mm, for tree={fill=yellow!20}
				[Infusing Embeddings, text width=3cm, for tree={fill=green!20}, minimum height=6mm, fill=yellow!20
					[
					IE \cite{IE},
					text width=2.2cm,
					node options={align=center}
					    [ConceptNet, 
					    text width=2.2cm, 
					    node options={align=left}, 
					    rounded corners=false,
					    fill=blue!5]
					]
					[
					MRG \cite{MRG},
					text width=2.2cm,
					node options={align=center}
					    [ConceptNet; self-constructed KG, 
					    text width=4cm, 
					    node options={align=left}, 
					    rounded corners=false,
					    fill=blue!5]
					]
					[
					SHGN \cite{SHGN},
					text width=2.2cm,
					node options={align=center}
					    [ConceptNet; self-constructed KG,
					    text width=4cm, 
					    node options={align=left}, 
					    rounded corners=false,
					    fill=blue!5]
					]
				]
				[Infusing Distribution, text width=3cm, for tree={fill=green!20}, minimum height=6mm, fill=yellow!20
					[
					GRF \cite{GRF},
					text width=2.2cm,
					node options={align=center}
					    [ConceptNet, 
					    text width=2.2cm, 
					    node options={align=left}, 
					    rounded corners=false,
					    fill=blue!5]
					]
					[
					MKR \cite{MKR},
					text width=2.2cm,
					node options={align=center}
					    [ConceptNet; COMET, 
					    text width=3cm, 
					    node options={align=left}, 
					    rounded corners=false,
					    fill=blue!5]
					]
					[
					CONPER \cite{CONPER},
					text width=2.2cm,
					node options={align=center}
					    [ConceptNet, 
					    text width=2.2cm, 
					    node options={align=left}, 
					    rounded corners=false,
					    fill=blue!5]
					]
					[
					KID \cite{liu2022KID},
					text width=2.2cm,
					node options={align=center}
					    [Wikipedia, 
					    text width=2.2cm, 
					    node options={align=left}, 
					    rounded corners=false,
					    fill=blue!5]
					]
				]
			]
		]
	\end{forest}
	\caption{A methodological taxonomy of methods on structured knowledge-enhanced story generation. Yellow boxes denote proposed (sub)categories. Green boxes denote existing works under their respective subcategories. Light blue boxes are the structured knowledge that each method utilizes.}
    \label{fig:taxonomy}
    \vspace{-0.3cm}
\end{figure*}

\textcolor{\revisioncolor}{Over the years, many Transformer-based autoregressive PLMs have been released, e.g., BART (Bidirectional and Auto-Regressive Transformers) \cite{Bart}, COMET (Commonsense Transformers) \cite{COMET}, and the GPT (Generative Pre-trained Transformer) series \cite{GPT-3,ouyang2022training}; and these can be readily used as building blocks for different generation tasks.}
Moreover, recent non-autoregressive language models~\cite{gu2018nonautoregressive} can generate word sequences in parallel without modeling sequential dependencies \cite{yang2021pos}.
Such models has been applied to neural machine translation and significantly speed up the generation process \cite{xiao2022survey}. However, the quality of their generated text is less satisfactory compared to autoregressive language models.

\subsection{Structured Knowledge}
Regarding the knowledge involved in story generation, there is both \textit{unstructured knowledge} and \textit{structured knowledge}.
Unstructured knowledge is implicit information inherent in text. 
Such knowledge is not shaped in a uniform form, but can be indicated and inferred from context.
By contrast, structured knowledge is explicit information shaped in uniform forms.
Many structured knowledge datasets have been constructed, such as triple-formed \cite{DBpedia,Freebase,ConpectNet,ATOMIC}, quadruple-formed \cite{temporalKG,GDELT}, tree-formed \cite{VerbNet,WordNet}, and table-formed \cite{wikidata} datasets.
Knowledge graphs are an example of triple-form datasets that collect a mass of real-world facts.
Each triple is usually in the form of (\textit{subject}, \textit{relation}, \textit{object}), where the subject and object entities refer to real-world objects or concepts, and the relation indicates the semantic relationship between the entities.
There are many large knowledge graphs \cite{Freebase,DBpedia,WordNet,Nell,YAGO3} storing fine-grained and instance-level knowledge like (\textit{Albert Einstein}, \textit{was born at}, \textit{Ulm}). 
Differently, commonsense knowledge graphs \cite{ConpectNet,ATOMIC} provide general rules in the everyday world that all humans are expected to know. 
Such rules can be commonsense facts like (\textit{lake}, \textit{contains}, \textit{water}), or causal relationships between events like (\textit{go for a run}, \textit{requires}, \textit{put on running shoes}).
Aside from story generation, structured knowledge has facilitated other popular tasks, like question answering \cite{ijcai2021p0611}, dialogue systems \cite{ni2022recent}, recommendation systems \cite{guo2020survey}, and code completion \cite{lu-etal-2022-reacc}.

\section{Integrating Structured Knowledge Into Story Generation: A Taxonomy} \label{sec_method}
We gather and analyze existing works on structured knowledge-enhanced story generation; and classify their methods into two strategic categories regarding how they integrate structured knowledge into the generation process: 
\begin{itemize}
    \item \textbf{\textcolor{\revisioncolor}{Structured Knowledge as Text}}. 
    Methods under this category view structured knowledge as materials that can be learned in the same way as stories.
    They transform structured knowledge into natural language text and feed them into generative models to capture their explicit information.
    This leads generative models to be inherently knowledge-enhanced.
    \item \textbf{\textcolor{\revisioncolor}{Structured Knowledge as Encoding}}.
    Methods under this category treat structured knowledge as an external instructor.
    They independently encode structured knowledge into low-dimensional vectors where the semantic correlation between concepts is captured. 
    These knowledge encodings are then used to influence word probability distribution over vocabulary from which stories are generated.
\end{itemize}

We give detailed descriptions of the two strategies next.
Fig.~\ref{fig:taxonomy} presents our taxonomy, where (sub)categories and their respective works are listed. 

\begin{figure*}[!t]
\centering
\includegraphics[width=0.9\textwidth]{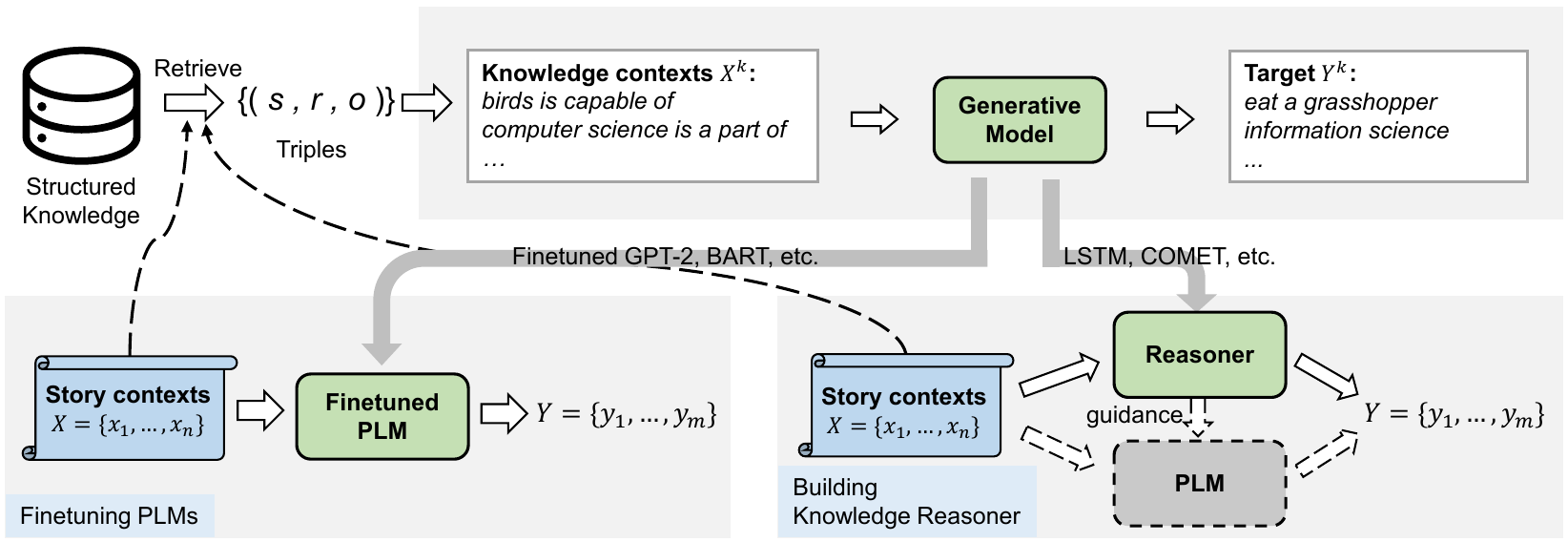}
\caption{Two strategies of knowledge as text category: \textit{finetuning PLMs} and \textit{building a knowledge reasoner}. ($s$, $r$, $o$) is the abbreviation for knowledge triple ($subject$, $relation$, $object$). The dotted box and arrows indicate optional processes.}
\label{fig:knowledge_as_text}
\vspace{-10pt}
\end{figure*}

\subsection{Structured Knowledge As Text} \label{sec_knowledge_as_text}
For this integration strategy, the first step is to transform structured knowledge data like knowledge triples into natural language text. 
This transformation can be realized through simple concatenation of elements in a triple \cite{martin2017improvisational,ammanabrolu2019guided,Coep} or pre-defined conversion templates \cite{Common-Posttrain,MEGATRON}.
The templates map each hard relation text into a vivid natural description, which we will discuss in Sec.~\ref{sec_operation}.
Each transformed natural language text is then divided into an input context and a predicted target.
The input context consists of the subject entity and the related elements of the knowledge triple, and the predicted target contains the object entity.
The input context is fed into a generative model to predict the target sequence, either by directly finetuning on PLMs, prompting, or training a knowledge generative model from scratch.
We refer to the first implementations as \textit{finetuning PLMs}, and the later implementation as \textit{building knowledge reasoner}.
Fig.~\ref{fig:knowledge_as_text} presents the common procedure of each.
Sections \ref{sec:finetuning_plms} and \ref{sec:building_reasoner} introduce their respective works and details.

\subsubsection{Finetuning PLMs}\label{sec:finetuning_plms}
The first work about knowledge-enhanced story generation is Knowledge-Enhanced GPT-2 (KEG) \cite{Common-Posttrain} proposed by Guan et al., which targets the starting sentence-prompted commonsense story generation task. 
It integrates the commonsense knowledge graphs ConceptNet \cite{ConpectNet} and ATOMIC \cite{ATOMIC} into GPT-2 \cite{GPT-2}, a large PLM.
They post-train GPT-2 with the triple-transformed knowledge texts that are generated by templates.
Formally, given a sequence of knowledge context $X^k =\{x_1^k, x_2^k, ..., x_q^k\}$ as input context, it is fed into GPT-2 to predict the target knowledge sequence $Y^k =\{y_1^k, y_2^k, ..., y_r^k\}$ through autoregressive language modeling.
The learning objective $\mathcal{L}_{KG}$ is:
\begin{equation} \label{eq:kg_loss}
    \mathcal{L}_{KG} = - \sum_{t=1}^{r} \log P(y_t^k | y_{<t}^k, X^k, \boldsymbol{\theta}),
\end{equation}
where $\boldsymbol{\theta}$ are the parameters of GPT-2.
The knowledge-enhanced GPT-2 is then finetuned on a story corpora by minimizing $\mathcal{L}_S$ Eq.~(\ref{eq:lm_loss}). 
To better distinguish true and fake stories, Guan et al. additionally train a binary classifier.
They shuffle, replace, and repeat sentences in true stories to create fake ones. 

Following KEG, Xu et al. proposed MEGATRON-CNTRL \cite{MEGATRON} which also feeds knowledge-transformed contexts into PLMs. 
But instead of first finetuning PLMs on knowledge text independently, it inputs knowledge and story contexts simultaneously to generate subsequences.
To be more specific, it first transforms all triples in ConceptNet into corresponding natural language sentences using templates. 
Next, it selects knowledge sentences related to the story.
To achieve this, it extracts keywords from the input context using the RAKE algorithm \cite{RAKE}. 
A knowledge retriever then uses the extracted keywords and queries ConceptNet to retrieve related knowledge sentences.
After obtaining these sentences, it further utilizes BERT \cite{Bert} to encode and calculate similarity between retrieved knowledge sentences and inputted story context.
The knowledge sentences with high similarity are kept and inputted into GPT-2, along with the story context, to generate the subsequent story.

COEP \cite{Coep} focuses on the commonsense logic when generating a future event. 
Given a current story event and its previous context, it frames the story generation process into two modules:
i) an inference module that reasons knowledge explanations out of input story contexts, such as intentions and causes, and ii) a generation module that generates future events prompted by the knowledge explanations.
In its inference module, COEP finetunes BART, a PLM, with commonsense knowledge from ATOMIC in the same way as KEG.
But different from transforming triples into text using pre-defined templates, COEP simply concatenates the subject entity and the relation of a triple to form the input knowledge context, then feeds it into BART to predict the target entity. 
The ATOMIC-enhanced BART can generate commonsense explanations for inputted story events.
Afterward, in the generation module, COEP finetunes another BART with knowledge from ConceptNet to better learn sequential dependency of events. 
The input story events packed with commonsense explanations are fed into the Concept-enhanced BART to predict future events.

\subsubsection{Building Knowledge Reasoners}\label{sec:building_reasoner}
Instead of directly generating discourse-level stories in an end-to-end manner by finetuning PLMs, methods that build knowledge reasoners dissect the generation process.
They train an independent knowledge reasoner from scratch using merely structured knowledge, which works as an event space.
Thus, these knowledge reasoners are able to infer on the plot level.
Then, the inferred plot lines participate in guiding discourse-level story generation.

Martin et al. \cite{martin2017improvisational,martin2018event,martin2018dungeons} proposed a pipeline where story generation can roughly be broken into two steps: \textit{event-to-event} and \textit{event-to-sentence}.
The event-to-event step generates successive events and the event-to-sequence step grounds an event to a discourse sentence. 
Both steps are trained with seq2seq models (e.g. RNN and LSTM).
Due to the sparsity of extracted fine-grained events from input texts, Martin et al. define each event as a 4-tuple representation $<s,v,o,m>$ where the nouns (subject $s$ and object $o$) and verb (verb $v$) are generalized using structured knowledge datasets WordNet \cite{WordNet} and VerbNet \cite{VerbNet}.
$m$ is a modifier of an event.
These tuples are inputted into a seq2seq model to predict successor events.
Then, another seq2seq model takes the predicted event as input and generates a sentence back to natural language discourse.
Following the pipeline, Ammanabrolu et al. \cite{ammanabrolu2019guided} explore an ensemble-based model in the event-to-sentence step to make generated sentences more on-topic and interesting.

Another knowledge reasoning model is COMET \cite{COMET}, which is built for commonsense knowledge completion.
COMET is a Transformer-based model trained with knowledge triples of ConceptNet and ATOMIC.
Given a seen pair consisting of a subject entity and a relation, such as (\textit{PersonX goes to the store}, \textit{xNeed}), COMET can complete the commonsense triple by inferring the object entity ``\textit{bring a wallet}''.
C2PO \cite{C2PO} utilizes COMET to simulate the inferential capability of humans. 
It targets the plot-infilling task where plot lines of a story are extracted and then need to be elaborated upon.
To tackle this, C2PO employs COMET to generate intermediate events between any two consecutive plot points. 
These events contribute to a plot graph that contains lots of possible event flows.
Then, C2PO walks a path from the starting event in the plot graph to the ending event.
It calculates the probability of event co-occurrence and chooses the most possible subsequent event.
After obtaining the elaborate events under the story plot lines, C2PO simply concatenates the events to form the final detailed story.

CAST \cite{CAST} aims at generating character-centered stories where each sentence should contain one or two characters.
It utilizes COMET as an inferential instructor to supervise the story generation process.
Specifically, given a story sentence as an input context, on the one hand, CAST employs GPT-2 to generate successive sequences.
On the other hand, CAST inputs the story sentence into COMET and obtains a commonsense inference explaining it, such as the character's intentions and desires.
The generated sequences must conform to the inferential explanations by COMET, otherwise re-generation is required.
In the confirmation stage, CAST encodes the generated sequences and the inferential explanations with sentence-BERT \cite{sentence-BERT} and calculates the cosine similarity between them to measure their likeness.
A whole character-centered story is generated by conducting this process iteratively.

\begin{figure*}[!t]
\centering
\includegraphics[width=0.9\textwidth]{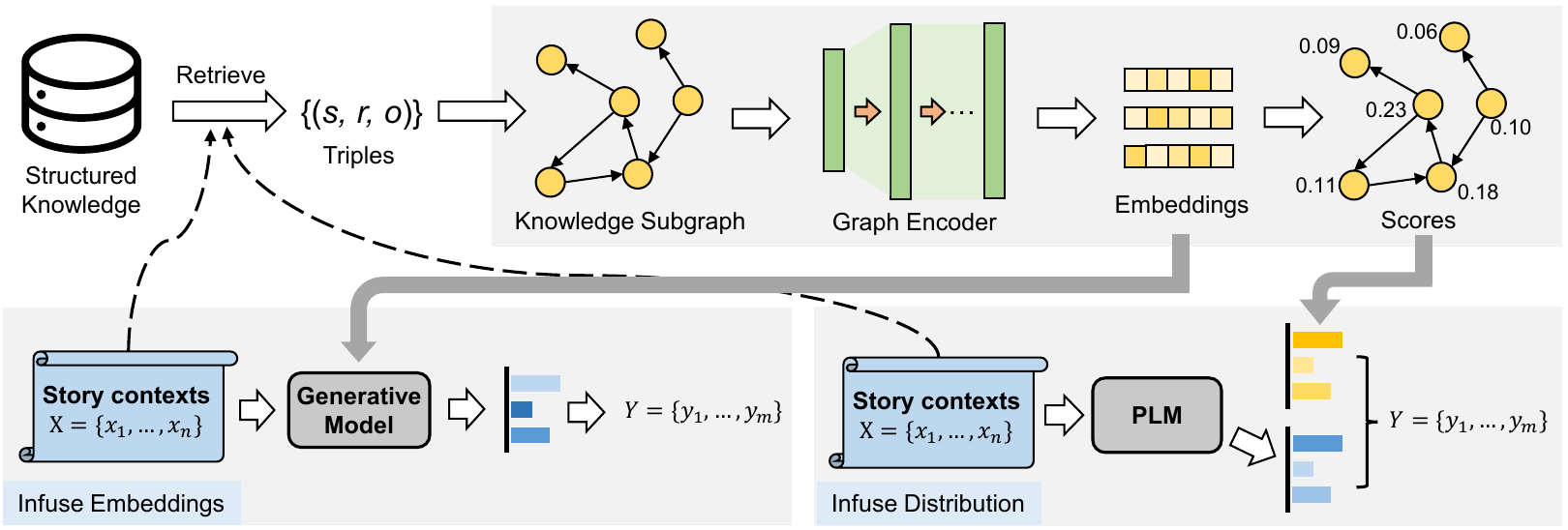}
\caption{Two strategies of the knowledge as encoding category: \textit{infusing embeddings} and \textit{infusing encodings}. ($s$, $r$, $o$) is the abbreviation for knowledge triple ($subject$, $relation$, $object$). The dotted box and arrows indicate optional processes.}
\label{fig:knowledge_as_encoding}
\vspace{-10pt}
\end{figure*}

\subsection{Structured Knowledge As Encoding} \label{sec_knowledge_as_encoding}
For methods that view knowledge as encoding, they intend to capture semantic information within original knowledge graphs.
Compared to converting knowledge triples into natural language sentences that can only reflect triple-level information, utilizing graph-structured knowledge can capture numerous triples within multiple hops at once.
Generally, these methods first retrieve knowledge subgraphs from a knowledge base and encode subgraphs into low-dimensional embeddings using graph encoders such as graph neural networks (GNNs).
Then, these embeddings are processed to twist the probability distribution over a vocabulary, so that semantic-relevant words from the knowledge graphs are more likely to be chosen.

GNNs are deep learning models for performing inference on graph data.
It consists of multiple stacked aggregation layers which iteratively update the embeddings of nodes (and edges).
Formally, given a knowledge graph $\mathcal{G}=\{\mathcal{E}, \mathcal{R}, \mathcal{T}\}$, where $\mathcal{E}$, $\mathcal{R}$, and $\mathcal{T}$ ($\mathcal{T} \subseteq \mathcal{E} \times \mathcal{R} \times \mathcal{E}$) denotes the set of entity, relation, and triple respectively,
GNNs update the embedding $\mathbf{h}_{e_i}^l$ of entity $e_i \in \mathcal{E}$ in $l$-th layer by aggregating its neighboring and self information:
\begin{equation}
\begin{split}
    \mathbf{h}_{e_i}^{l} = \sigma \Bigl( & \frac{1}{|\mathcal{N}(e_i)|} \sum_{(e_j, r)\in\mathcal{N}(i)} \mathbf{W}^l_{N} \phi\bigl(\mathbf{h}_{e_j}^{l-1}, \mathbf{h}_r^{l-1}\bigr) \\ 
    & + \mathbf{W}^l_S \mathbf{h}_{e_i}^{l-1}\Bigr).
\end{split}
\end{equation}
Here, $\mathcal{N}(e_i)$ denotes the set of directly neighboring entity-relation pairs of $e_i$.
$\mathbf{h}_{e_i}$, $\mathbf{h}_{e_j}$, and $\mathbf{h}_{r}$ are latent embeddings.
$\mathbf{W}_{N}$ and $\mathbf{W}_{S}$ are two learnable weight matrices.
$\phi(\cdot)$ is a transition function that reflects the information passed from specific neighbors. 
And $\sigma(\cdot)$ is an activation function.
Iterative information propagation on graphs can capture semantic relationships among entities multi-hop away.

Regarding how the output embeddings of the last layer of graph encoders can be used to twist the final word distribution and influence sequence generation, we further divide it into two implementations: \textit{infusing embeddings} and \textit{infusing distributions}.
For the former, the embeddings are injected into manipulable generative models such as LSTMs as internal parameters.
The subgraph embedding-infused generative model then directly generates subsequent sequences through the language model.
The whole process happens as a black box.
Differently, the implementation of infusing distributions calculates a score for each entity in the subgraph based on these embeddings.
The scores measure the relevance between subgraph entities and the input story context. 
Later, the scores are infused with the probability distribution generated by PLMs over vocabulary as the final distribution.
An overall illustration of knowledge as encoding category is presented in Fig.~\ref{fig:knowledge_as_encoding}.

\subsubsection{Infusing Embeddings}
Methods following this implementation integrate structured knowledge at the embedding level.
Since manipulating the internal parameters of PLMs can cause disastrous degradation in performance \cite{Plug-Play}, this implementation does not adopt PLMs as their generative models but train seq2seq models like LSTMs and Transformers from scratch.
IE \cite{IE} focuses on the story ending generation task, which aims at generating an ending sentence for a story body.
Given the input context, a one-hop knowledge subgraph for each conceptual word in the context is retrieved from ConceptNet.
Each subgraph is then encoded by a GNN. 
The output embeddings of conceptual words are generated in the last layer.
These embeddings later are fed into an LSTM network together with conceptual words' contextual embeddings calculated from their story context.
The LSTM network tries to predict the target sentence.

MRG \cite{MRG} focuses on long text generation which necessitates content planning beforehand.
It constructs a domain-specific knowledge graph from input materials using Point-wise Information \cite{PMI}.
In the graph, each entity corresponds to a word appearing in the input texts.
Entities are connected if their corresponding words appear in the same context.
Also, there are weights on the edges which are calculated by word co-occurrence. 
These weights indicate the relevant possibility two words can co-occur.
Based on the self-constructed knowledge graph, MRG conducts multi-hop reasoning and seeks possible paths as planned content.
Starting from an entity, it walks to the next neighboring entity in the graph until meets an ending entity.
Specifically, for each visited entity, MRG predicts whether to discard, continue, or finish this walk using a BERT-based labeler.
After this, each path is represented as the sum of embeddings of all entities in the path. 
The path embedding worked as planned content is then fed into a Transformer together with story context to generate subsequent sequences.

Wang et al. \cite{SHGN} also target the story ending generation task.
They propose a model named SHGN (Story Heterogeneous Graph Network) to capture both the information within the story context and the information retrieved from structured knowledge.
The constructed heterogeneous graph contains three types of entities: context keywords, context sentences, and one-hop neighboring concepts of context keywords retrieved from ConceptNet.
They utilize SimCSE \cite{SimCSE}, a large-scale pre-trained embedding model, to generate initial embeddings for graph entities.
Then, they input the initialized embeddings into a heterogeneous graph Transformer \cite{HGT} to capture the semantic relationship among entities.
The final output embeddings of the heterogeneous graph Transformer are passed through Transformer decoders to generate story endings.

\newcolumntype{b}{>{\hsize=1.\hsize}X}
\newcolumntype{m}{>{\hsize=.9\hsize}X}
\newcolumntype{o}{>{\hsize=.55\hsize}X}
\newcolumntype{s}{>{\hsize=.5\hsize}X}
\newcolumntype{a}{>{\hsize=.32\hsize}X}

\begin{table*}[!t]
    \centering
    \caption{\textcolor{\revisioncolor}{Summary table of the features and techniques used in structured knowledge-enhanced story generation works. SSPG, SEG, and PI represent starting sentence-prompted generation, story ending generation, and plot infilling, respectively. The Generative Model and Generation Level columns of KID are left blank as KID is an algorithm that can be applied to various models at different generation levels.}}
    \footnotesize
        \begin{tabularx}{\linewidth}{l c m a b s o}
            \toprule
            Year & Work  & Knowledge Integration \newline Strategy & Task \newline Scenario & Task Focus & Generative \newline Model  & Generation \newline Level \\
            \toprule
            2017 & Martin et al. \cite{martin2017improvisational,martin2018event,martin2018dungeons} & knowledge as text & SSPG & story event generalization & LSTM  & plot, discourse \\
            2019 & Ammanabrolu et al. \cite{ammanabrolu2019guided} & knowledge as text & SSPG & story event grounding  & LSTM & plot, discourse \\
            2019 & IE \cite{IE} & knowledge as encoding & SEG & commonsense reasoning  & LSTM& discourse \\
            2020 & KEG \cite{Common-Posttrain} & knowledge as text & SSPG & knowledge awareness & GPT-2 & discourse \\
            2020 & MEGATRON-CNTRL \cite{MEGATRON} & knowledge as text & SSPG & story controllability & GPT-2 & discourse \\
            2020 & MRG \cite{MRG} & knowledge as encoding & SSPG & long story generation & Transformer & plot, discourse \\
            2020 & GRF \cite{GRF} & knowledge as encoding & SEG & commonsense reasoning & GPT-2 & discourse \\
            2021 & C2PO \cite{C2PO} & knowledge as text & PI & commonsense reasoning & COMET & plot, discourse \\
            2021 & CAST \cite{CAST} & knowledge as text & SSPG & character-centric narrative & COMET & plot, discourse \\
            2022 & COEP \cite{Coep}& knowledge as text & SSPG & commonsense reasoning & BART & plot, discourse \\
            2022 & CONPER \cite{CONPER} & knowledge as encoding & SSPG & persona awareness & GPT-2 & discourse \\
            2022 & SHGN \cite{SHGN} & knowledge as encoding & SEG & commonsense reasoning & Transformer & discourse \\
            2022 & MKR \cite{MKR} & knowledge as encoding & SEG & commonsense reasoning & COMET, BART & discourse \\
            2022 & KID \cite{liu2022KID} & knowledge as encoding & SEG & knowledge awareness & --- & --- \\
            \bottomrule
        \end{tabularx}
    \label{table:sum_up}
\end{table*}

\subsubsection{Infusing Distribution}
Rather than directly infusing subgraph embeddings into generative models, this strategy infuses structured knowledge at the final level of story generation, where sequences of words are selected from a probability distribution.
Each entity in the knowledge subgraph is calculated with a relevance score using embeddings.
The scores reflect the relevant probability for the corresponding word over the subgraph being selected at the last stage.
By combining the distribution generated by PLMs and probability distribution over words in the subgraph, this strategy twists the selection process towards knowledge words highly relevant to the input story context.

GRF \cite{GRF} conducts multi-hop reasoning on commonsense knowledge subgraphs.
Given an input story context, GRF recognizes concepts from the input and retrieves their neighboring concepts from ConceptNet.
Then, GRF utilizes a GNN to encode entities and relations in the subgraph into low-dimensional embeddings.
These embeddings participate in propagation to calculate a probability score for each concept.
Specifically, GRF first assigns a score of 1 to each concept appearing in the input. 
Then, for other concepts, GRF updates their scores hop by hop using an embedding-based scoring function.
In the last, all calculated scores are re-scaled to $(0,1)$ by $\mathsf{softmax}$ as the local probability over concepts in the subgraph.
This probability distribution is then combined with the probability distribution generated by GPT-2 over vocabulary using a gate mechanism. 
The subsequent sequences are then generated from the combined word distribution.

Based on GRF, MKR \cite{MKR} argues that the word-level knowledge graph like ConceptNet sometimes cannot capture the semantics of events precisely.
It constructs a multi-level knowledge graph, including an event-level subgraph and a word-level subgraph.
To be more specific, it first extracts events from the input story context and uses COMET to generate subsequent events.
All these events contribute to an event-level subgraph.
Then, a word-level subgraph is constructed by dissecting each event into dispersed words which are connected by ConceptNet relations.
After this, MKR encodes the multi-level knowledge graph using a GNN.
The learned embeddings are used for calculating probability scores for words in the word-level subgraph in the same way as GRF.
Finally, the probability distribution over the subgraph is combined with the distribution generated by BART to generate sequential sequences. 

CONPER \cite{CONPER} targets the story generation scenario where the generated texts should reflect the persona of the main characters.
Given a story context and a piece of personality description of the main characters in the story, CONPER first extracts keywords from the story context and personality description that have an emotional tendency to accord with main characters' persona. 
Then, it extends these keywords by retrieving neighboring words from ConceptNet and constructs a local knowledge subgraph.
Likewise, words in the knowledge subgraph are encoded into embeddings by a GNN and later fed into a full convolutional layer with $\mathsf{softmax}$ in the end to calculate the probability distribution. 

Very recently, Liu et al. propose a knowledge infused decoding (KID) \cite{liu2022KID} algorithm for generative PLMs. They dynamically infuse external knowledge, which are Wikipedia word chunks, into each step of the PLM decoding. Specifically, for each story context, this work first retrieves relevant knowledge entities from a dynamically maintained knowledge memory. Then, in each step, it calculates a knowledge gain from the probabilities over the retrieved knowledge and uses the knowledge gain to define a reward that the PLM is finetuned to maximize. In a nutshell, KID reshapes the probability distribution at each decoding step toward the entities in knowledge.

\subsection{Summing Up}
\textcolor{\revisioncolor}{Here, we recapitulate the key features of the discussed works, encompassing their knowledge integration strategies, task scenarios, task focuses, generative models used, and the generation levels of stories.
We present the summary in Table~\ref{table:sum_up} and arrange these works chronologically to provide a better view of the evolution in the field.}

\textcolor{\revisioncolor}{
It is easy to note that, while many of these works share a common strategy in integrating structured knowledge, they display notable diversity in terms of task scenarios and focuses.
For those targeting the starting sentence-prompted generation scenario, the earliest works focus on generalizing and grounding story events by leveraging structured knowledge to produce fluent and coherent stories.  However, later works shift their attention towards specific attributes of the generated text, such as knowledge awareness, the ability to generate longer narratives, and enhancing controllability of content.
For works targeting story ending generation and plot infilling, they predominantly emphasize incorporating commonsense reasoning when generating text, as both of these scenarios rely heavily on causal relationships among events.
Moreover, it is evident that, over time, large PLMs have become the favored backend choice as generative models.
Experimental evidence showcases their advantage in capturing language patterns due to their training on vast and diverse corpora of text data (and sometimes non-textual data), outperforming non-pretrained language models significantly \cite{Common-Posttrain,MEGATRON}. 
Furthermore, plot-level generation is essential for works that build knowledge reasoners into their integration strategy and those focusing on long story generation.
Such intermediate level products offer overarching guidance and increase the controllability for long narrative text generation.
Overall, the choice of backend model and design of integration strategies in structured knowledge-enhanced story generation works remains highly task-dependent.
}
\section{Knowledge-related Operations} \label{sec_operation}
Some methods described above require non-trivial operations such as extracting keywords and transforming triples into natural language sentences.
These operations are critical for the effectiveness of utilizing structured knowledge.
For methods that are under the knowledge as text strategy, they need to convert knowledge triples into readable texts before processing them into generative models.
For methods that utilize structured knowledge as encoding, they need to firstly extract knowledge entities from inputted story context, and then build a local knowledge subgraph.
In the following, we introduce the implementations used in existing works and other practicable approaches.

\subsection{Transforming Triples to Sentences}
The simplest implementation is concatenating a triple's entities and relation. 
The input context in the training of COMET is basically doing so.
Moreover, COMET adds a ``\textit{[MASK]}'' token between the entity and relation to make them distinguishable.
For example, given the triple (\textit{natural language understanding, part of, artificial intelligence}) from ConceptNet, COMET converts to ``\textit{natural language understanding [MASK] part of [MASK] artificial intelligence}''. 
But direct concatenation can introduce ``out-of-bag'' tokens to process due to the special schema of knowledge datasets, such as the relation ``\textit{xIntent}'' in \textsc{Atomic}.
Using templates can avoid such nuisance because templates convert unique tokens to readable natural language while expressing the same meaning.
Levy et al. \cite{Levy} provide a pre-defined set that maps each relational type $R(x, y)$ in knowledge data to at least one readable natural language sentence.
For example, the relational type $educated\_at(x, y)$ can be transformed into the sentence ``\textit{x graduated from y}'', where $x$ and $y$ are the subject entity and object entity.
Other works \cite{cimiano2013exploiting,CORAL,turner-etal-2009-generating} also provide triple-to-text rules or templates.
Aside from templates, many neural network-based works \cite{VOUGIOUKLIS20181,yang-etal-2020-improving-text,Triple-to-Text} translate knowledge triples into natural language sentences by training a seq2seq model.

\subsection{Extracting Keywords, Keyphrases, and Events}
One naive way to extract keywords is checking if a word's corresponding entity exists in a structured knowledge dataset. 
However, checking words one by one is time-consuming, especially when retrieving from vast datasets. 
Although some words appear in the knowledge dataset, they are trivial or unrelated to the topic of the input story. 
Including them can introduce noises to retrieved knowledge.
To avoid these problems, IE \cite{IE} and GRF \cite{GRF} only consider nouns and verbs as the candidate keywords and check them on ConceptNet.
MEGATRON-CNTRL \cite{MEGATRON} utilizes \texttt{RAKE} algorithm \cite{RAKE} to extract keywords from the input context.
\texttt{RAKE} is an unsupervised and domain-independent algorithm for extracting keywords from text.
But it can still cause information loss for sometimes it cannot extract all keywords successfully \cite{MEGATRON}.
CONPER \cite{CONPER} extracts emotional keywords from the persona description of protagonists by utilizing the \texttt{NLTK} sentiment analyzer.
If the emotional tendency of a word is strong ``\textit{positive}'' or ``\textit{negative}'', this word is considered as a keyword.
As for extracting events, C2PO \cite{C2PO} and CAST \cite{CAST} use the tool \texttt{OpenIE} \cite{OpenIE} to extract relational triples as events and use a pre-trained neural coreference resolution model \cite{Coreference} to find all character coreference.
With the extracted events and character coreference, the plot lines can be built upon.
Martin et al. \cite{martin2018event} use Stanford's \texttt{CoreNLP} \cite{manning-etal-2014-stanford} to extract word dependencies that form fine-grained events. 
Then they generalize the fine-grained events by finding hypernyms in WordNet and VerbNet. 
In addition to these approaches, there are many statistical and neural network-based extraction methods for keywords \cite{Matsuo,Cohen,Luhn,ramos2003using}, keyphrases \cite{Campos2020YAKEKE,bennani-smires-etal-2018-simple,liang-etal-2021-unsupervised,ye-etal-2021-heterogeneous}, and events \cite{okamoto2009discovering,liu2008extracting,wang_event,zhou-etal-2017-event}.
Since statistical approaches usually require no manually annotated training data, they are convenient to use for diverse domains \cite{bharti2017automatic}.

\subsection{Building Knowledge Subgraphs}
There are different ways to build local knowledge subgraphs based on extracted keywords.
IE \cite{IE} and CONPER \cite{CONPER} simply retrieve one-hop neighbors of keywords from ConceptNet to form local graphs.
MRG \cite{MRG} needs to find paths for content planning. 
Thus, it explores multi-hop neighbors in ConceptNet or self-constructed graphs with a maximum hop constraint to avoid over-exploration.
GRF \cite{GRF} also explores multiple hops in ConceptNet, but only keeps salient concepts that are frequently visited to support information flow.
Wang et al. \cite{SHGN} only retain concepts that are connected with keywords in more than one context sentence to control the quality of retrieved concepts.
MKR \cite{MKR} utilizes COMET to infer multiple subsequent events for extracted key events.
The key events and inferred events build an event-level local graph.
Then, MKR dissects events into words.
These words together with their relational connections constitute ConceptNet form a word-level local graph.

\begin{table*}[!b]
    \centering
    \caption{Detailed statistics and characteristics of collected story corpora for story generation. Symbol ``\#'' stands for number. ``Drive'' and ``Causality'' stands for the drive of story and causality between events, respectively.}
    \footnotesize
    \begin{threeparttable}
        \resizebox{\textwidth}{!}{
        \begin{tabular}{l l c c c c c c}
             \toprule
             \multirow{2}{*}[-3pt]{\textbf{Story Corpora}} & \multirow{2}{*}[-3pt]{\textbf{Source}} & \multirow{2}{*}[-3pt]{\textbf{\# Story}} & \multirow{2}{*}[-3pt]{\textbf{\shortstack{\# Sentence\\Per Story}}} & \multirow{2}{*}[-3pt]{\textbf{\shortstack{\# Word\\Per Story}}}  & \multicolumn{3}{c}{\textbf{Characteristics}} \\
             \cmidrule{6-8}
                      &     &     &   &   &  \textbf{Genre}  & \textbf{Drive} & \textbf{Causality}  \\
             \toprule
             
             \textsc{Children’s Book Test}  \cite{CBT}  & Ebooks\tnote{1}  & $687,343$ &  $21$ & $464.7$ & fairy tale  & character-driven &  \\ 
             
             \multicolumn{7}{l}{~~~$\circ$ \url{https://research.facebook.com/downloads/babi}} \\

            \midrule
             
             \textsc{Visual Storytelling} \cite{visual_storytelling} & Flickr\tnote{2}  & $50,200$  &  $5$  & $56.7$  &  realistic fiction  & plot-driven & \checkmark \\ 
             
             \multicolumn{7}{l}{~~~$\circ$ \url{http://visionandlanguage.net/VIST}} \\

             \midrule
             
             \textsc{ROCStories}  \cite{ROCStories}   &  AMT writers\tnote{3}  &  $98,161$ &  $5$  & $49.8$ &  realistic fiction &  plot-driven  &  \checkmark \\
             
             \multicolumn{7}{l}{~~~$\circ$ \url{https://cs.rochester.edu/nlp/rocstories}} \\

             \midrule
             
             \textsc{Writing Prompts} \cite{HNN} & Reddit\tnote{4}  & $303,358$  & $45.7$  & $738.2$  & fiction & plot-driven &  \\
             
             \multicolumn{7}{l}{~~~$\circ$ \url{https://www.kaggle.com/datasets/ratthachat/writing-prompts}} \\

             \midrule
             
             \textsc{Fairy Tales} \cite{FairyTale_Mistery}   & Wikipedia  & $850$ &  $21.5$  & $543.4$ & fairy tale &  character-driven &   \\
             \multicolumn{7}{l}{~~~$\circ$ \url{https://github.com/rajammanabrolu/WorldGeneration}} \\

             \midrule
             
             \textsc{Mystery}  \cite{FairyTale_Mistery}  & Wikipedia  & $532$ &  $17.6$  & $479.4$  & mystery & plot-driven & \\
             
             \multicolumn{7}{l}{~~~$\circ$ \url{https://github.com/rajammanabrolu/WorldGeneration}} \\

             \midrule
             
             \textsc{STORIUM} \cite{STORIUM}  & Storium\tnote{5} &  $5,743$ & $1,145.4$ & $19,141.5$ & fiction & plot-driven & \\ 
             
             \multicolumn{7}{l}{~~~$\circ$ \url{https://storium.cs.umass.edu}} \\

             \midrule
             
             \textsc{Hippocorpus} \cite{Hippocorpus} & AMT writers\tnote{3} & $6,854$ & $17.6$ & $292.6$ & realistic fiction & character-driven & \checkmark \\ 
             
             \multicolumn{7}{l}{~~~$\circ$ \url{http://aka.ms/hippocorpus}} \\

             \midrule
             
             \textsc{TVRecap} \cite{TVRecap} & Fansites\tnote{6} & $29,013$ & $9.0$ & $147.8$ & fiction & plot-driven & \\ 
             
             \multicolumn{7}{l}{~~~$\circ$ \url{https://github.com/mingdachen/TVRecap}} \\

             \midrule
             
             \textsc{LiSCU} \cite{LiSCU} & Study guides\tnote{7} & $1,708$ & $48.8$ & $1,022.3$ & fiction & character-driven & \\ 
             
             \multicolumn{7}{l}{~~~$\circ$ \url{https://github.com/fabrahman/char-centric-story}} \\

             \midrule
             
             \textsc{STORAL-EN-MO2ST} \cite{STORAL} & Websites\tnote{8} & $1,779$ & $17.7$ & $302.3$ & fiction & plot-driven & \checkmark \\ 
             
             \multicolumn{7}{l}{~~~$\circ$ \url{https://github.com/thu-coai/MoralStory}} \\

             \midrule
             
             \textsc{STORAL-ZH-MO2ST} \cite{STORAL} & Websites\tnote{8} & $4,209$ & $17.6$ & $321.8$ & fiction & plot-driven & \checkmark \\ 
             
             \multicolumn{7}{l}{~~~$\circ$ \url{https://github.com/thu-coai/MoralStory}} \\
             
            \bottomrule
        \end{tabular}}
        
        \begin{tablenotes}
            \item[1] Ebooks from Project Gutenberg (\url{https://www.gutenberg.org}).
            \item[2] Flickr (\url{https://www.flickr.com}) is an image and video hosting service.
            \item[3] Human writers from Amazon Mechanical Turk (\url{https://www.mturk.com}).
            \item[4] The \texttt{r/WritingPrompts} community (\url{https://www.reddit.com/r/WritingPrompts/}) on Reddit.
            \item[5] Storium (\url{https://storium.com}) is an online creative writing game. 
            \item[6] Fan-contributed websites including Fandom (\url{https://www.fandom.com/}) and TVMegaSite (\url{http://tvmegasite.net/}).
            \item[7] Study websites such as Shmoop (\url{https://www.shmoop.com/}) which contain educational literature material.
            \item[8] A list of websites such as Aesop (\url{https://www.read.gov/aesop/}) which contain moral stories.
        \end{tablenotes}
    \end{threeparttable}
    \label{table:story_datasets}
\end{table*}
\newcolumntype{B}{>{\hsize=1.4\hsize}X}
\newcolumntype{s}{>{\hsize=.9\hsize}X}
\newcolumntype{n}{>{\hsize=.5\hsize}X}
\newcolumntype{t}{>{\hsize=.3\hsize}X}

\begin{table*}[!t]
    \centering
    \caption{Detailed statistics and genres of triple-formed knowledge datasets utilized in story generation. Symbol `\#' stands for the number. ``Event'' denotes the entities in role of base events in \textsc{ATOMIC} and \textsc{ATOMIC}2020.}
    \footnotesize
        \begin{tabularx}{\linewidth}{l s l n t l B}
            \toprule
            \textbf{Knowledge Dataset} & \textbf{Source} & \textbf{\# Triple} & \textbf{\# Entity \newline (Event)} & \textbf{\# Relation} & \textbf{Genre} & \textbf{Instance} \\  
            \toprule
            \textsc{ConceptNet} (v.5) \cite{ConpectNet} & OMCS & $34,074,917$ &  $28,370,083$  & $50$ & general & (\textit{cat}, \textit{is capable of}, \textit{hunt mice}), \newline (\textit{book}, \textit{has}, \textit{knowledge}), \newline (\textit{water}, \textit{is used for}, \textit{drink}) \\

            \midrule
            
            \textsc{Atomic}  \cite{ATOMIC}  & Stories, Books, \newline Google Ngrams, \newline Wiktionary Idioms & $877,108$ &  $309,515$ \newline ($24,313$) & $9$ & inferential &  (\textit{PersonX adopts a dog}, \textit{xEffect}, \textit{PersonX then smiles}), \newline (\textit{PersonX smiles at PersonY}, \textit{oReact}, \textit{PersonY feel happy}) \\

            \midrule
            
            \textsc{Atomic-2020}  \cite{ATOMIC_2020}  & Crowd-sourcing, \newline \textsc{ATOMIC}, \newline \textsc{ConceptNet} & $1,331,113$ & $638,127$ \newline ($43,958$) & $23$ & inferential & (\textit{X gets X’s car repaired}, \textit{happens after}, \textit{X drives an old car}), \newline (\textit{money}, \textit{has property}, \textit{earned by working})\\

            \midrule
            
            $\textsc{GLUCOSE}$  \cite{GLUCOSE} & \textsc{ROCStories} & $304,099$ & $306,903$  & $10$ & inferential & (\textit{Times passes}, \textit{causes}, \textit{Someone\_A grows}), \newline (\textit{It is winter}, \textit{enables}, \textit{It is cold}) \\

            \midrule
            
            $\textsc{TransOMCS}$ \cite{TransOMCS}  & \textsc{ASER} & $18,481,607$ & $100,659$ & $20$ &  linguistic & (\textit{student}, \textit{AtLocation}, \textit{school}), \newline (\textit{hilarious}, \textit{InstanceOf}, \textit{mess}) \\

            \midrule
            
            $\textsc{TupleKB}$ \cite{Dalvi2017}  & Elementary Science & $282,594$ & $6,465$ & $1,605$ & scientific & (\textit{moose}, \textit{eat}, \textit{plant}), \newline (\textit{engine}, \textit{use}, \textit{energy}) \\
            
            \bottomrule
        \end{tabularx}
    \label{table:knowledge_dataset}
\end{table*}

\section{Datasets} \label{sec_datasets}
In this section, we collect available story corpora and structured knowledge datasets for story generation.
We present their statistics and characters from a practical view.

\subsection{Story Corpora}
Many story corpora covering different genres and characteristics have been constructed so far.
\textsc{Children's Book Test} dataset \cite{CBT} is originally built to measure how well language models can understand the wider linguistic context.
It collects stories from 98 online children's books written by different writers.
In the last sentence of each story, one word is removed for the language models to answer multiple choices.
\textsc{Visual Storytelling} dataset \cite{visual_storytelling} is the first dataset for the sequential vision-to-language task.
It contains sequences of images where each image is described with an event and sequential images constitute a coherent story.
\textsc{ROCStories} dataset \cite{ROCStories} is a collection of five-sentence commonsense stories written by human writers.
It is a widely-used benchmark for story generation for the conciseness and generality of the stories. 
Also, each story conveys clear causal relation between its constituted events.
\textsc{Writing Prompts} dataset \cite{HNN} collects coherent and fluent passages of human-written online stories. 
These stories are prompted by given topics or cues.
\textsc{Fairy Tales} and \textsc{Mystery} datasets \cite{FairyTale_Mistery} contain the main plots of fairy tales and mystery short stories. 
They are extracted from the Wikipedia page.
\textsc{STORIUM} dataset \cite{STORIUM} collects stories from a collaborative online storytelling community.
Its stories are lengthy and fine-grained with structural annotations.
\textsc{Hippocorpus} dataset \cite{Hippocorpus} contains diary-like short stories of recalled and imagined events.
These stories are also written by human writers.
\textsc{TVRecap} dataset \cite{TVRecap} collects detailed TV show episode recaps from fan-contributed websites.
Its stories feature complex interactions among multiple characters.
\textsc{LiSCU} dataset \cite{LiSCU} is created from various online study guides.
It contains literary pieces and their summaries paired with descriptions of characters that appear in them.
\textsc{STORAL} \cite{STORAL} is a collection of moral stories each of which is paired with a moral sentence. 
These moral stories are extracted from multiple online websites.
It provides both Chinese and English moral stories.

We present the sources, statistics, and characteristics of these story datasets in Table~\ref{table:story_datasets}.
We download and calculate their total story number, average sentence number per story, and average word number per story. 
For the latter two figures, we calculate using \texttt{sent\_tokenize} and \texttt{word\_tokenize} of the \texttt{nltk} package.
Specifically, for \textsc{Fairy tales} and \textsc{mystery} datasets, since their original data can not be found in their GitHub repository, we re-scrap the latest stories from the Wikipedia page using their codes and calculate their statistics.
For \textsc{Storal} dataset, we present the statistics of its English and Chinese stories constructed for the moral-to-story generation task, which is about generating a story given a moral sentence as a prompt.
The last three columns are the characteristics of these datasets. 
We summarize the characteristics from three practical dimensions: (i) genre of story; (ii) drive of story; (iii) causality between events.
We consider these dimensions important for choosing training corpora for a specific task.
We label each story corpus with a genre like ``fairy tale'' or ``realistic fiction''.
The ``realistic fiction'' means the stories could have actually occurred in real life.
Meanwhile, a story can be driven by characters or plots.
Character-driven stories mainly focus on the characters themselves, including their thoughts and conversations.
While plot-driven stories focus on events, including action, cause, and effect.
Furthermore, if the stories generally express strong causality between events, this corpus is marked with ``$\checkmark$''. 
We also provide their download addresses.

\subsection{Structured Knowledge Datasets}
The external structured knowledge used for enhancing story generation is mostly triple-formed commonsense knowledge datasets such as \textsc{ConceptNet} and \textsc{Atomic}.
The reasons why they are preferred are that, on one hand, compared to fine-grained knowledge, commonsense knowledge introduces fewer off-topic noises for a given story context.
On the other hand, commonsense knowledge like \textsc{Atomic} explicitly indicates the causality between events which increases the logical coherence of stories.
\textsc{ConceptNet} originated from the crowd-sourcing project OMCS (Open Mind Common Sense), with more knowledge sources such as DBpedia \cite{DBpedia} and Wiktionary being merged over time.
Each entity in \textsc{ConceptNet} denotes a commonsense concept such as ``\textit{water}'' and ``\textit{drink}'', and each relation indicates the semantic relation between two concepts, such as ``\textit{is used for}''.
The concepts are mostly words of different parts of speech, while some others are phrases.
\textsc{Atomic} is an atlas containing inferential knowledge of generic events.
Each entity in \textsc{Atomic} is a textual description of an event or a conclusion following an event which indicates the character's mental state or persona. 
The relations in \textsc{Atomic} consist of $9$ kinds of ``\textit{if-event-then}'' inferential relations between two entities.
\textsc{Atomic-2020} \cite{ATOMIC_2020} enlarges \textsc{Atomic} by incorporating social and physical aspects of everyday inferential knowledge, as well as absorbing more inferential knowledge of events.
The relation set of \textsc{Atomic-2020} expands to $23$ different commonsense relations, including $9$ relations of social interaction, $7$ physical-entity relations, and $7$ event-centered relations.
\textsc{GLUCOSE} \cite{GLUCOSE} extracts 10 dimensions of causal explanation from sentences in \textsc{ROCStories}, and generalizes them as semi-structured inference rules.
Compared to \textsc{Atomic} which contains non-contextual knowledge, \textsc{GLUCOSE} grounds inferential knowledge to a particular story context.
Furthermore, \textsc{GLUCOSE} extends \textsc{Atomic}'s person-centric knowledge to wider topics such as places and things.
\textsc{TransOMCS} \cite{TransOMCS} is a knowledge graph that collects high-quality commonsense knowledge from the linguistic graph ASER \cite{zhang2020aser}, whose nodes are words and edges are linguistic relations.
The TupleKB \cite{Dalvi2017} dataset contains high-precision, domain-specific knowledge tuples extracted from scientific text using an extraction pipeline, and guided by domain vocabulary constraints.
In Table~\ref{table:knowledge_dataset} we present the statistics of these triple-formed commonsense knowledge graphs and their respective commonsense genre.

Aside from commonsense knowledge graphs in triple-form, tree-formed datasets such as WordNet \cite{WordNet} and VerbNet \cite{VerbNet} are used in works \cite{martin2017improvisational,martin2018dungeons,martin2018event,ammanabrolu2019guided} to generalize events.
WordNet is a large lexical database of English.
It gathers synonyms of words and creates super-subordinate relations between most synonym sets such as ``ISA'' and ``hyponymy''.
Synonym sets of adjectives are interlinked in terms of antonymy. 
Totally WordNet created around $117,000$ synonym sets. 
VerbNet \cite{VerbNet} is also a hierarchical network of English words but focuses only on verbs. 
It maps WordNet, PropBank \cite{PropBank}, and FrameNet \cite{FrameNet} verb types to their corresponding verb classes.
Both WordNet and VerbNet can easily be read using \texttt{NLTK} interfaces. 

\section{Evaluation Metrics} \label{sec_eval}
In this section, we summarize involved evaluation metrics in story generation and discuss their specialties. 
We classify these evaluation metrics into three categories: \textit{statistical}, \textit{neural network-based}, and \textit{manual}.
The summaries of metrics under each category are listed in Table~\ref{table:evaluation}.

\begin{table*}[!ht]
    \caption{The list of evaluation categories and specific metrics for story generation. Arrow direction in performance indicates whether higher or lower the metric score is considered better.}
    \footnotesize
    \begin{tabularx}{\linewidth}{l c c X}
         \toprule
         \textbf{Metric} & \textbf{Category} & \textbf{Performance} & \textbf{Description} \\
         \midrule
         Perplexity \cite{perplexity} & Statistical & $\downarrow$ & Inverse probability of the target sentences. \\
         Repetition-$n$ \cite{repetition} & Statistical &  $\downarrow$ & Measure the redundancy of stories. The ratio of repeated $n$-grams to all $n$-grams. \\
         Distinct-$n$ \cite{li2016diversity}  & Statistical & $\uparrow$ & Measure the diversity of stories. The ratio of distinct $n$-grams to all $n$-grams. \\
         Coverage \cite{Common-Posttrain} & Statistical &  $\uparrow$ & The amount of commonsense triples that are covered in generated stories. \\
         BLEU-$n$ \cite{BLEU} & Statistical   &  $\uparrow$ & Evaluate $n$-gram overlap between two pieces of texts. \\
         ROUGE \cite{ROUGE}  & Statistical    &  $\uparrow$ & A metric package used for evaluating the similarity between two pieces of texts.\\
         METEOR \cite{METEOR}  & Statistical   &  $\uparrow$ &  Measure the similarity between two pieces of texts. It has features like stemming and synonymy matching. \\
         CIDEr \cite{CIDEr}  & Statistical     &  $\uparrow$ & A metric measuring image descriptions using human consensus. \\
         
         BertScore \cite{BERTScore} & NN-based  &  $\uparrow$ & Evaluate semantic similarity between two pieces of texts which are encoded by Bert. \\
         UNION  \cite{UNION}  & NN-based   &  $\uparrow$ & An unreferenced metric for evaluating open-world story generation. \\
         MAUVE \cite{MAUVE} & NN-based   &  $\uparrow$ & Measure the gap between two pieces of texts using divergence frontiers. \\
         
         Grammar \cite{Common-Posttrain} & Manual &  $\uparrow$ & Indicate whether a story reads natural and fluent. \\
         Logicality \cite{Common-Posttrain} & Manual  &  $\uparrow$ & Indicate whether a story is coherent to the given beginning and reasonable in terms of causal and temporal dependencies in the context. \\
         Consistency \cite{MEGATRON} & Manual  &  $\uparrow$ & \textcolor{\revisioncolor}{Indicate} whether a story follows the same topic from beginning to end. \\
         Informativeness \cite{MKR} & Manual    &  $\uparrow$ & \textcolor{\revisioncolor}{Assess} whether the generated text produces unique and non-genetic information that is specific to the input context. \\  
         Interestingness \cite{CAST} & Manual   &  $\uparrow$ & Indicate whether the story is enjoyable or not. \\    
        \bottomrule
    \end{tabularx}
    \label{table:evaluation}
\end{table*}

\subsection{Statistical Evaluation Metrics}
\noindent\textbf{Perplexity:}
The perplexity (PPL) \cite{perplexity} metric measures how well a language model predicts a target sequence.
For a target sequence $Y=\{y_1,y_2,...,y_m\}$, the PPL score is calculated as:
\begin{equation}\label{eq:PPL}
    \mathrm{PPL}(Y) = \frac{1}{\sqrt[m]{p(y_1)p(y_2) \cdots p(y_m)}},
\end{equation}
\begin{equation}
    p(y_i) = \frac{\mathrm{count}(y_i)}{\sum_{y_j \in Y} \mathrm{count}(y_j)}.
\end{equation}
Here, $\mathrm{count}(\cdot)$ is the frequency of occurrence of a word. 
PPL can be interpreted as the inverse probability of a target sequence normalized by the number of words.
A smaller PPL score indicates better prediction performance.
But for story generation, a model's confidence is hardly equal to this metric due to the diversity of gold discourse and stories. 
Besides, this metric is not task-specific and can not measure the model's confidence on the event level.

\noindent\textbf{Repetition-$n$ and Distinct-$n$:}
The redundancy and diversity of generated stories can be measured with Repetition-$n$ \cite{repetition} and Distinct-$n$ \cite{li2016diversity}, with $n$ is usually set to $4$. 
Repetition-$n$ calculates the percentage of tokens that repeat at $n$-gram, while distinct-$n$ measures the ratio of distinct $n$-grams in generated tokens. 
The lower repetition-$n$ and higher distinct-$n$ indicate better diversity of stories. 
However, these two metrics also focus on word-level repetition/distinction and are unable to notice event-level and plot-level similarity/difference.

\noindent\textbf{Coverage:}
The coverage metric \cite{Common-Posttrain} measures the model on how well it utilizes external knowledge. 
It calculates the average number of knowledge triples that appeared in each generated story. 
A triple is considered utilized only if its head and tail entities both appear in the generated story.

\noindent\textbf{BLEU-$n$:}
BLEU-$n$ (BiLingual Evaluation Understudy) \cite{BLEU} evaluates $n$-gram overlap between a generated story and a target story. 
Let $\hat{Y}$ and $\hat{l}$ denote the generated story sequence and its token length, and $Y$ and $l$ denote the target sequence and its token length.
BLEU-$n$ is computed as follows:
\begin{equation}\label{eq:BLEU-n}
    \textrm{BLEU} = BP \times \exp\biggl(\sum_{i = 1}^{n}w_i \log p_i\big(\hat{Y}, Y\big)\biggr),
\end{equation}
\begin{equation}\label{eq:BLEU-BP}
    BP = 
    \begin{cases}
    1,                       & \text{if } l_c > l_r~; \\ 
    e^{1-l_r/l_c},   & \text{if } l_c \leq l_r~.
    \end{cases}
\end{equation}
Here, $p_i\big(\hat{Y}, Y\big)$, which is valued in $[0,1]$, is the precision measuring the normalized number of $i$-gram in $\hat{Y}$ appearing in $Y$.
$w_i$ is the weight on $i$-gram precision. In most cases, all weights are set to $1/n$.
$BP$ is a penalty to punish generated sequences that are too short. 

In story generation, $n$ is usually set to $1$ or $2$.
Caccia et al. \cite{Massimo2020language} find that BLEU-$n$ failed to catch the semantic or global coherence in long generated text, which results in stories with poor information content but with correct grammar that can also get a perfect BLEU score.

\noindent\textbf{ROUGE:}
Compared with BLEU-$n$ which calculates overlap between two pieces of sequences, ROUGE (Recall-Oriented Understudy for Gisting Evaluation) \cite{ROUGE} pays more attention to how much information of the target sequence can be recalled in the generated sequence. 
It calculates the normalized number of $n$-grams in the target sequence that appears in the generated sequence, where $n$ is set to $1$ or $2$.

Story generation has the notorious one-to-many issue, which means there can be plenty of good sequential sequences for the same story context but only one target sequence is given. 
Thus, BLEU-$n$ and ROUGE are both limited for open-world story generation \cite{HNN}.

\noindent\textbf{METEOR:}
The METEOR \cite{METEOR} improves BLEU-$n$ by using WordNet for stemming and synonym matching $n$-grams. 
Besides, METEOR computes both precision and information recall.
However, due to its utilization of WordNet, METEOR can not evaluate generated sequences containing words outside WordNet's lexicon. 
And it is also inadequate for story generation, for there can be plenty of good sequential sequences aside from the target one.

\noindent\textbf{CIDEr:}
CIDEr (Consensus-based Image Description Evaluation) \cite{CIDEr} is originally proposed to evaluate image captions. Compared with the above $n$-gram matching evaluation metrics, CIDEr \cite{CIDEr} further uses tf-idf (Term Frequency Inverse Document Frequency) to weight different $n$-gram when calculating similarity. 
Therefore, CIDEr is able to focus on ``important'' words in sequences.

\subsection{Neural Network-based Evaluation Metrics}
With the emergence of deep neural networks, some neural network-based evaluation methods are proposed to lessen the gap between automatic evaluation with human judgments. 

\noindent\textbf{BERTScore:}
Unlike previous methods that calculate token-level \textcolor{\revisioncolor}{syntactic} similarity, BERTScore \cite{BERTScore} computes the similarity between the generated sequence and the target sequence using their contextual embeddings generated by BERT. 
The scores are calculated with pairwise cosine similarity between two embeddings. 
BERTScore is much more flexible in matching two sequences' semantics. 
However, the grammar errors might be ignored due to the comparison between only embedding not words.

\noindent\textbf{UNION:}
UNION \cite{UNION} is a Bert-based metric that does not need a target sequence for reference to measure the generated story.
It is trained as a classifier aiming to distinguish human-written stories from negative samples.
These negative samples are created with manipulations including repetition, substitution, reordering, and negation alteration.
Thus, UNION has the ability to capture commonly observed issues like repeated plots and conflicting logic in generated stories.

\noindent\textbf{MAUVE:}
MAUVE \cite{MAUVE} compares the learned distribution from a language model to the distribution of human-written texts using divergence frontiers.
It can reflect the quality of generated stories affected by length, decoding algorithm, and model size.
These properties are not captured and analyzed in existing metrics.

\subsection{Manual Evaluation Metrics}
Though manual evaluation is costly but still necessary for evaluating generated stories due to the limitations of automatic metrics \cite{see-etal-2019-massively}.
A service for manual evaluation is Amazon Mechanical Turk\footnote{\url{https://www.mturk.com/}} (AMT), where annotators are asked questions covering the following dimensions of story quality and give a score on each dimension.

\noindent\textbf{Grammar:}
Measuring the grammar correctness of generated texts such as subject-verb agreement, tense, singular/plural agreement, punctuation, and sentence fragments.

\noindent\textbf{Logicality:}
This measures whether the events, characters' actions, and intentions in generated stories accord with the input contexts and are reasonable.
A good story should act up to audiences' inference.
However, the logicality is hard to measure for automatic metrics.

\noindent\textbf{Consistency:}
This measuring whether the generated stories and given context are on the same topic and under persistent rules.
Some inconsistencies in a story are like protagonists' ``out of character'', the settings behind stories contradict before and after.

\noindent\textbf{Informativeness:}
This dimension reflects how much knowledge users can perceive when reading the generated stories.
The coverage metric is designed for a similar purpose by counting knowledge triples appearance without being aware of knowledge relevance. 
Given two pieces of generated sequences, they may contain the same number of knowledge triples, but their informativeness can vary much due to different choices of triples.

\noindent\textbf{Interestingness:}
Whether a story is interesting is hard to evaluate but essential to readers. 
A story might be reasonable and fluent but uninteresting.
Nowadays, studies just let human judges distinguish whether a story is interesting or not according to their tastes, which can vary from individual to individual. 
Maybe there's still a long way to go to measure the story's interestingness automatically.

\section{Discussion on Challenges} \label{sec_limitation}
Incorporating structured knowledge has significantly advanced open-world story generation, but the quality of generated stories is still far from human-written counterparts.
To bridge the gap, efforts are required to tackle multi-aspect challenges that hinder the way.
The challenges include the inherent limitation of PLMs, the information loss when utilizing structured knowledge, and even the evaluation metrics.
In the following, we dissect each challenge and discuss solutions for them.

\subsection{Limitation of PLMs for Storytelling}
Although PLMs have demonstrated remarkable text generation capabilities, the generated text has shown the prevalence of being generic, including the scope and precision of word choosing \cite{gretz-etal-2020-workweek}. 
This is due to the training materials of PLMs, which are crowd-sourced from broad genres while falling short of reflecting the characteristics of story narration.
Overcoming this inherent limitation of PLMs is challenging, as it incurs significant costs to train or fine-tune the models \cite{GPT-3}.
In many research domains, such as bio-medicine, engineering, and finance, people began to train domain-specific PLMs on respective resources \cite{gu2021domain, ZHENG2022103733, yang2020finbert}, and these models have brought great advances in each field.
Likewise, we can train storytelling-specific PLMs exclusively on story corpora to make the generated narratives more story resembled.
Though some existing works like MRG \cite{MRG} and SHGN \cite{SHGN} train a Transformer decoder from scratch, they are corpus-oriented and thus produce stories with certain narration styles.
However, to realize storytelling-specific PLMs, training on large-scale and diverse corpora is necessary.

\subsection{Limitation of Knowledge Sources}
Another impediment is the limited structured knowledge sources. 
Almost all existing works on structured knowledge-enhanced story generation only utilize commonsense knowledge datasets like \textsc{ConceptNet} and \textsc{Atomic}.
Although commonsense knowledge inherits abundant generic facts and inferential explanations, which can boost stories' logical coherence and increase their readability, making stories more informative and descriptive can be tricky.
For example, imagine generating a story about a real person such as \textit{Emily Dickinson}; if the model can be aware of Dickinson's family background, occupation, and literary works, the generated story can be more convincing and vivid.
Unfortunately, commonsense knowledge datasets do not contain such detailed facts to elaborate on.
Also, PLMs show unsatisfactory performance in expressing fine-grained knowledge because they are trained with online texts which overgeneralize or contain big noises for specific knowledge.

Incorporating fine-grained knowledge is challenging because such knowledge datasets, like DBpedia \cite{DBpedia} and Wikidata \cite{wikidata}, are enormous.
Each entity usually links many varied neighboring entities and attributes.
This causes retrieving information from these knowledge datasets to introduce irrelevant noise from the given story context.
Besides, since fine-grained knowledge is not generic and needs to cater to specific contexts, it requires real-time retrieval during generation instead of being finished while preprocessing.
Such real-time retrieval from vast databases can increase generation time costs.
Overall, a trade-off is necessary when incorporating fine-grained knowledge. 

\subsection{Information Loss in Knowledge-related Operations}
During the intermediate generation process, non-trivial operations such as keyword extraction and triple-to-text conversion can cause information loss, hindering knowledge enhancement's desired performance boost.
Many methods \cite{Common-Posttrain, MEGATRON,Coep} transform knowledge triples into texts by pre-defined templates.
While such templates can yield readable texts, they are limited in linguistic and creative diversity.
For each relation, templates only paraphrase it to one or a few natural language phrases, which is far more monotonous than how diversely and vividly meaning can be conveyed in the real world.
Because language models generate words according to the word correlation pattern in training data, template-based transformation may have difficulty sensing and elaborating upon knowledge when encountering a hint expressed outside pre-defined templates.
Thus, more vivid, diverse, and robust approaches to transforming knowledge triples into readable texts are needed.

Extracting keywords or key events from a story context to retrieve relevant knowledge can also involve information loss and noise.
If all words appearing in knowledge datasets are extracted, some of them will introduce off-topic information because not every word is a keyword.
Guan et al. \cite{Common-Posttrain} and GRF \cite{GRF} tackle this by only extracting nouns and verbs.
Meanwhile, extracting key events can produce inaccurate results and causes a detrimental effect on performance.
Therefore, more reliable extraction tools and algorithms are necessary to ensure the accuracy of extracted information.

\subsection{Limitation on Controllability}
Since PLMs produce token sequences from learned language models, their generated stories accord with the distribution of crowd-sourced data and usually deviate from readers' expectations \cite{StoRM,Alabdulkarim-Survey}.
Readers may deem such unexpectedness as creative because they are unfamiliar with the generated samples.
However, PLMs, in many circumstances, can be seen to mimic what they see in the training data \cite{Zellers2019,jones2022you,uchendu-etal-2020-authorship}.
As for readers' expectations, after they input a starting sentence, a list of keywords, or the plot lines of a story, in many cases, they have anticipations for story continuation, such as specific actions and endings.
How to control generative models to produce stories satisfying readers' expectations is trending in story generation studies.
Enhancement by structured knowledge is one of the manners of controllability, for it encourages the generated stories to reflect relevant knowledge to the input context.
It helps to keep the stories on topic and coherent.
Nevertheless, reflecting on relevant knowledge is still not enough.
Some studies try to make hard control on PLMs to produce guided words or plots from users' input.
Dathathri et al. \cite{Plug-Play} combine a PLM with a simple attribute classifier consisting of a user-specified bag of words.
This classifier alters the latent representations of input words and pushes the PLM's hidden activation to guide the generation.
PlotMachine \cite{PlotMachine} transforms a plot outline into a coherent story by tracking the dynamic plot states.
It loosely integrates key plots into the output narrative.
Pascua et al. \cite{Keyword2Text} add a shift to the probability distribution over vocabulary towards given guided words.
It modifies the log probability of the words according to their semantic similarity to the guide words, quantified as the cosine similarity between their vector representations.
Soft control aims at directing the mood or high-level topic of stories, which is more difficult but sensible.
Some studies \cite{CONPER,EmoRL} try to regulate the personalities or emotions of protagonists of stories.
However, more kinds of controllability are left to explore.

\subsection{Limitation on Commonsense Inference}
Commonsense inference in the text is about finding the relatedness between narrative events and their causal relationship. Considering two events ``\textit{win a lottery}'' and ``\textit{\textcolor{\revisioncolor}{feel excited}}'', one can easily infer that the former event is likely to ``\textit{result in}'' the latter. Conveying a clear inference between story events is critical for good readability and coherence.
Many works chose to finetune PLMs on inferential datasets like \textsc{ATOMIC} to achieve better inference awareness in generated stories.
However, as Liu et al. highlight in their work \cite{li2022textcis}, finetuning PLMs directly on inferential datasets can mess up both the language generation process and the commonsense inference process. They propose a new task named Context Commonsense Inference in Sentence Selection ($\text{CIS}^2$) to examine models' ability on commonsense inference better. This task abstracts away language generation and forces the language model to focus exclusively on inference.
But still, presenting awareness of commonsense inference is an open and challenging problem in story generation and other generation tasks \cite{xiang2022}.

\subsection{Limitations of Automatic Evaluation Metrics}
Another important issue for story generation is the lack of robust machine evaluation metrics.
Although many automatic evaluation metrics listed in Table~\ref{table:evaluation} have been used in existing works, they cannot accurately reflect the quality of generated stories.
These evaluations correlate poorly with human judgments \cite{liu-etal-2016-evaluate,sagarkar-etal-2018-quality}.
These metrics either score the similarity between generated stories with ``\textcolor{\revisioncolor}{golden answers}" (Perplexity, BLEU-$n$, ROUGE, METEOR, CIDEr, BertScore, MAUVE), measure how many knowledge triples appear in generated stories (Coverage), or check verbal repetition and diversity (Repetition-$n$, Distinct-$n$).
However, as is well-known, there is no \textcolor{\revisioncolor}{golden answer} to what is a good story, and perhaps there are infinite narratives to make a story good \cite{mccabe1984makes}.
Relying solely on one evaluation metric may underestimate interesting and good-quality stories. 
Thus, recruiting human judges to rate stories is indispensable for fair evaluation, despite the cost.
Recently, some hybrid-metric systems like ADEM \cite{lowe-etal-2017-towards}, BLEURT \cite{BLEURT}, and STORIUM \cite{STORIUM} aim to evaluate stories more comprehensively.
Nonetheless, more reliable evaluation metrics are needed.
One inspiration can be imitating humans' mental processes \cite{chang1983mental,Raymond}.

Guan et al. propose a benchmark OpenMEVA \cite{OpenMEVA} for automatic story generation metrics.
OpenMEVA provides a comprehensive test suite to assess the capabilities of different metrics.
They find that existing metrics all have a low correlation with human judgments and difficulty noticing off-topic and conflicting plots.
Xie et al. propose an approach named DeltaScore that utilizes perturbation to evaluate fine-grained story aspects \cite{xie2023deltascore}.
Their approach demonstrates a strong correlation with human evaluations.
\section{Future Directions} \label{sec_prospect}
\subsection{Non-text Storytelling}
Open-world storytelling can be conveyed through various mediums.
Recently, some compelling pre-trained generative models can create realistic images and art from a description in natural language. Examples include DALLE-1 \cite{DALLE-1}, DALLE-2 \cite{DALLE-2}, and Stable Diffusion \cite{stable-diffusion}. 
Incorporating text story generation with text-to-image generative models can present creative and vivid stories.
An interesting application of this approach is creating illustrations for generated stories, also known as a ``storyboard.''
However, existing pragmatic storyboard systems are mostly based on retrieving images for story text \cite{Picturing_Engine,Chowdhury,zakraoui2019text,Storyboard}. 
They are short of creative and flexible storytelling ability.
In 2023, Chen et al. propose a story generation pipeline for visual storytelling \cite{chen2023visual} that includes a narrative generation section and an image generation section.
At around the same time, Hong et al. proposed a character-based visual storytelling model driven by coherence \cite{hong2023visual}.

\subsection{Different Narrative Techniques}
Most open-world story generation works settle on chronological storytelling, where plots are presented in temporal order.
However, evidence has suggested that different discourse organizations of underlying events can affect readers' mental processes while reading and comprehending \cite{brewer1982stories}. Non-chronological narrative techniques like flashbacks and flashforwards can stir readers' curiosity.
Targeting this, Han et al. \cite{flashback} propose a hierarchical model that generates flashbacks in stories by encoding temporal dependencies of events as  temporal prompts and inputting them into the generative model.
Other narrative techniques like foreshadowing, plot twists, and multiple points of view deserve more attention for application.

\subsection{Collaborative Storytelling}
Most story generation tasks are about completing a story given an existing context, and little human-machine collaboration occurs except for inputting prompts from time to time.
However, collaborative storytelling enables humans and machines to mutually assist in their creativity \cite{goldfarb-tarrant-etal-2019-plan}. 
Several works have made efforts to incorporate more human creativity in automatic storytelling.
Goldfarb-Tarrant et al. \cite{goldfarb-tarrant-etal-2019-plan} propose an interactive system that allows flexible human involvement in story planning, editing, and revising without regeneration of the whole story.
Brahman et al. \cite{brahman-etal-2020-cue} focus on the scenario that users constantly provide cue phrases to drive the generation process. 
Other works such as Dramatron \cite{dramatron} achieve collaborative storytelling in specific domains like screenplays and theater scripts \cite{dirik2021controlled, HorrorWriter}. 
More sophisticated collaborations, such as recognizing discourse-level causal relations and character emotions, need further study.

\subsection{Generating Long Stories}
Existing story generation practices mostly target very short stories, which can be noticed from Table~\ref{table:story_datasets}. The stories used in their works are often less than $1,000$ words.
While in the real world, the stories humans read are actually much longer, and the limitation on length is a major obstacle for automatic story generation to be in a more extensive application. 
Generating long stories is difficult. The story needs to maintain a coherent throughout plot, preserve the same narrative style, and avoid inconsistency.
Recently, Yang et al. \cite{yang2022} propose a Recursive Reprompting and Revision framework to generate long stories.
As the first work to explicitly target long story generation, this work leaves promising room for further improvement and exploration, including a more abundant outline plan and more comprehensive inconsistency detection. 

\section{Conclusion} \label{sec_conclusion}
In this survey, we discuss the progress in the field of structured knowledge-enhanced story generation.
As the main contribution of this survey, we discuss the technical details of related works and present a \textcolor{\revisioncolor}{systematic} taxonomy regarding how they incorporate structured knowledge into story generation process: \textit{structured knowledge as text} and \textit{structured knowledge as encoding}.
\textcolor{\revisioncolor}{Through a comparison of the features of different works, we call attention to the diverse scenarios and focuses of approaches to story generation tasks.
The integration strategies and model designs of the analyzed works are highly task-dependent, making it unfeasible to put them under the same evaluation metrics.}
We summarize the characteristics of story corpora, knowledge datasets, and their evaluation metrics as contributions to foster further research. 
\textcolor{\revisioncolor}{Likewise, we raise insights into the current challenges of open-world story generation and cast light on promising directions for future study, especially on
what external knowledge sources to use and how to better leverage such knowledge.}
We hope this survey aid readers in understanding the latest developments in this field and inspire new research avenues.

\bibliographystyle{elsarticle-num}
\bibliography{ref}  

\end{document}